\documentclass{article}

\usepackage{PRIMEarxiv}

\usepackage[utf8]{inputenc} 
\usepackage[T1]{fontenc}    
\usepackage{hyperref}       
\usepackage{url}            
\usepackage{booktabs}       
\usepackage{amsfonts}       
\usepackage{nicefrac}       
\usepackage{microtype}      
\usepackage{lipsum}
\usepackage{fancyhdr}       
\usepackage{graphicx}       
\graphicspath{{media/}}     
\usepackage{amsmath}

\title{EchoVideo: Identity-Preserving Human Video Generation by Multimodal Feature Fusion
}

\author{
  Jiangchuan Wei \\
  ByteDance \\
  \texttt{weijiangchuan@bytedance.com} \\
   \And
  Shiyue Yan \\
  ByteDance \\
  \texttt{yanshiyue@bytedance.com} \\
  \And
  Wenfeng Lin \\
 ByteDance \\
 \texttt{linwenfeng.1008@bytedance.com} \\
  \And
  Boyuan Liu \\
 ByteDance \\
 \texttt{liuboyuan@bytedance.com} \\
  \And
  Renjie Chen \\
 ByteDance \\
 \texttt{chenrenjie.1998@bytedance.com} \\
  \And
  Mingyu Guo \\
  ByteDance \\
  \texttt{guomingyu.313@bytedance.com} \\
}

\begin{document}
\maketitle

\begin{figure}[h]
    \centering
    \vspace{-15mm}
    \includegraphics[width=\linewidth]{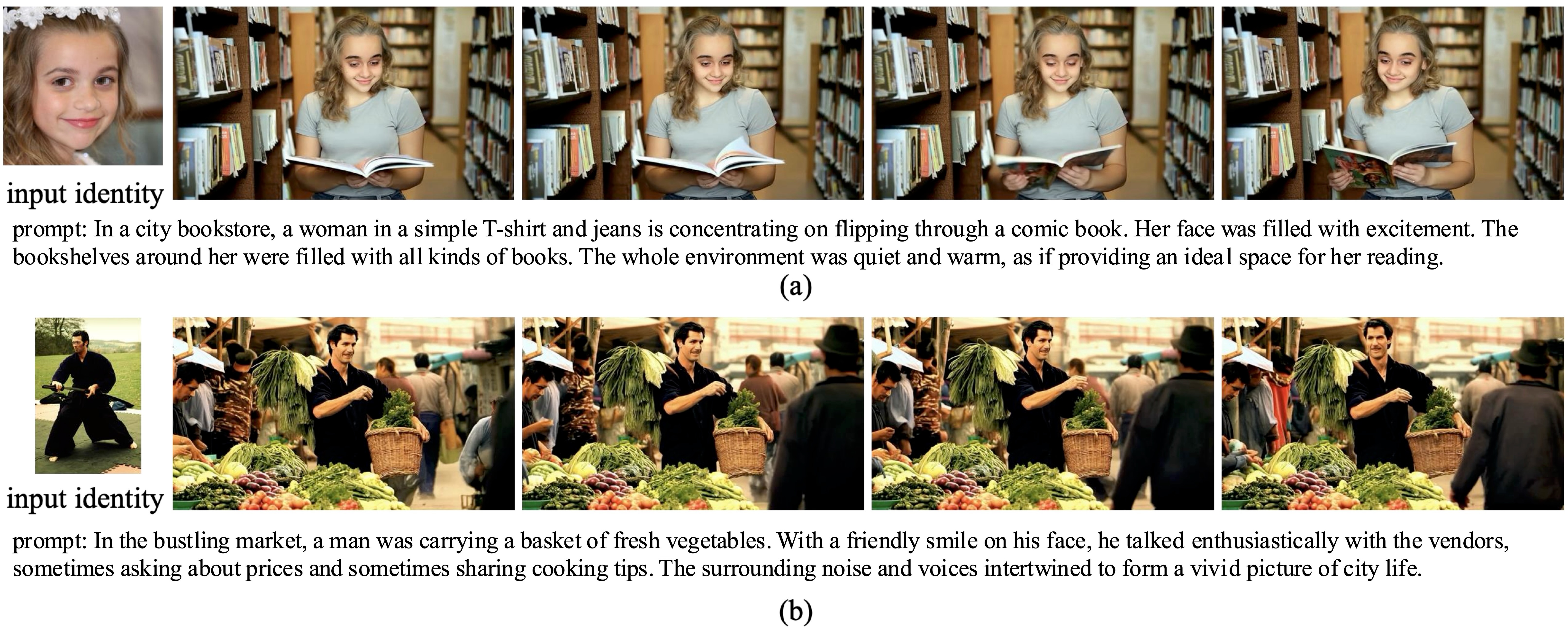}
    \caption{Sampling results of EchoVideo. (a) Facial feature preservation. (b) Full-body feature preservation. EchoVideo is capable of not only extracting human features but also resolving semantic conflicts between these features and the prompt, thereby generating coherent and consistent videos.}
    \label{fig:0}
\end{figure}

\begin{abstract}
   Recent advancements in video generation have significantly impacted various downstream applications, particularly in identity-preserving video generation (IPT2V). However, existing methods struggle with "copy-paste" artifacts and low similarity issues, primarily due to their reliance on low-level facial image information. This dependence can result in rigid facial appearances and artifacts reflecting irrelevant details. To address these challenges, we propose EchoVideo, which employs two key strategies: (1) an Identity Image-Text Fusion Module (IITF) that integrates high-level semantic features from text, capturing clean facial identity representations while discarding occlusions, poses, and lighting variations to avoid the introduction of artifacts; (2) a two-stage training strategy, incorporating a stochastic method in the second phase to randomly utilize shallow facial information. The objective is to balance the enhancements in fidelity provided by shallow features while mitigating excessive reliance on them. This strategy encourages the model to utilize high-level features during training, ultimately fostering a more robust representation of facial identities. EchoVideo effectively preserves facial identities and maintains full-body integrity. Extensive experiments demonstrate that it achieves excellent results in generating high-quality, controllability and fidelity videos. The code and model are available at: \url{https://github.com/bytedance/EchoVideo}.
\end{abstract}

\begin{figure}
    \centering
    \includegraphics[width=\linewidth]{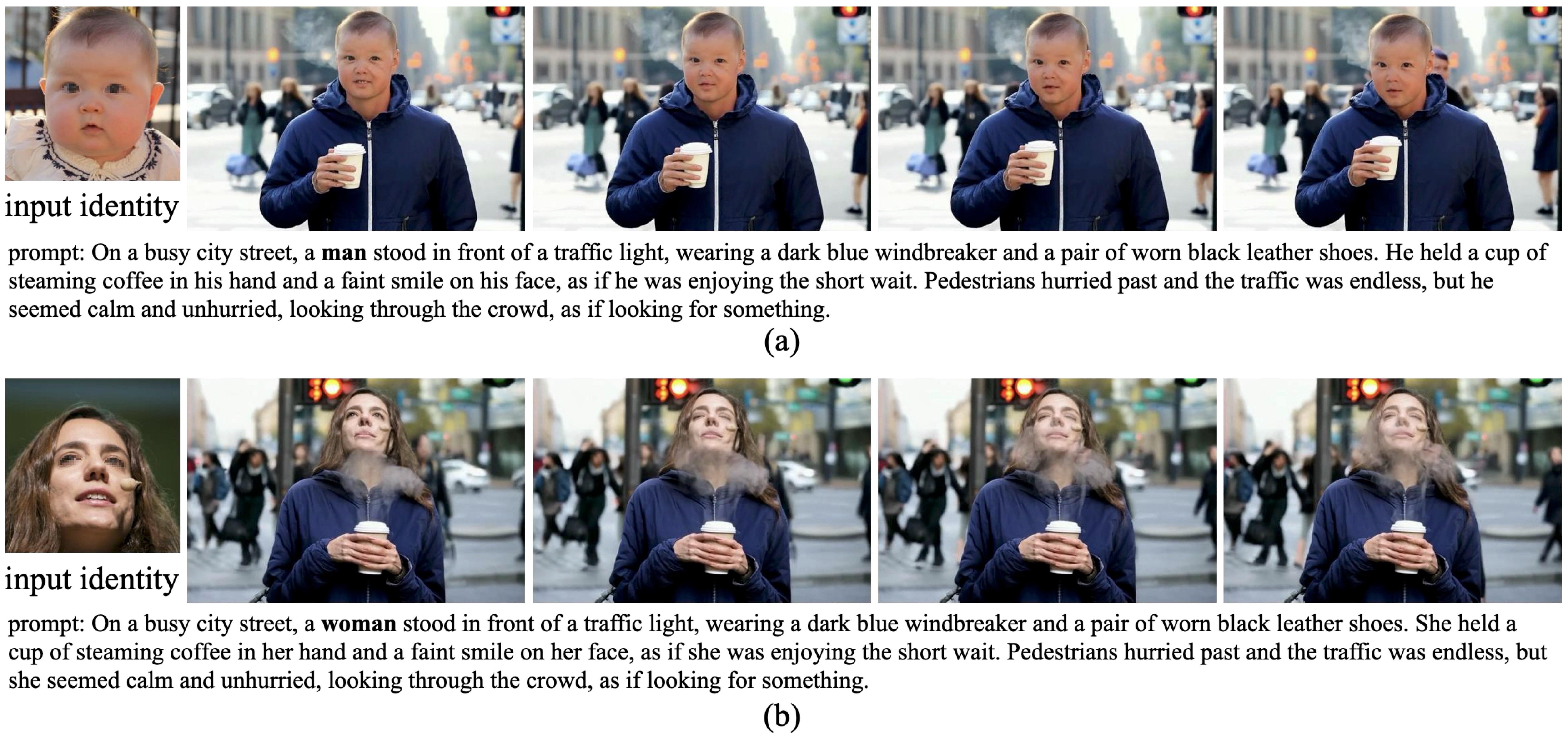}
    \caption{Issues in IP character generation. (a) Semantic conflict. The input image depicts a child's face, while the prompt specifies an adult male. Insufficient information interaction leads to inconsistent character traits in the model's output. (b) Copy-paste. During training, the model overly relies on visual information from facial images, directly using the Variational Autoencoder(VAE)-encoded \cite{yang2020causalvae} face as the output for the generated face.}
    \label{fig:1}
\end{figure}

\section{Introduction}

In the realm of text-to-video generation tasks, large-scale pre-trained models based on Diffusion Transformers (DiT)  \cite{peebles2023scalable} have exhibited exceptional performance, driving significant advancements across various downstream applications \cite{zhang2023adding,yu2024evagaussians,tang2024cycle3d,shi2024instadrag}, particularly in identity-preserving text-to-video (IPT2V) generation. Users can input images containing facial portraits, which are subsequently utilized by text-to-image models to produce videos that feature their likeness. Recently, several methods have emerged within the open-source community \cite{ma2024magic,he2024id,zhang2025magic,yuan2024identity} targeting this domain. Notably, ConsisID \cite{yuan2024identity} has achieved state-of-the-art (SOTA) verifiable results; However, it relies heavily on shallow information extracted from facial images, leading to artifacts such as rigid facial expressions and misalignment between the face and body in the generated outputs. Additionally, the generated faces are vulnerable to interference from extraneous information present in the input facial images, including occlusions. This issue is often associated with the "copy-paste" phenomenon. In contrast, closed-source methods such as Vidu \cite{shengshu2024vidu}, Pika \cite{pi2025pika}, and HailuoAI \cite{minimax2025hailuo} frequently encounter challenges related to low similarity between the generated faces and the input facial images.

In light of the observed phenomena, we note that the process of decoupling facial identity information from portrait images often incorporates excessive irrelevant information inherent to the images themselves. For instance, the ConsisID \cite{yuan2024identity} utilizes the original portrait images as strong supervisory signals, concatenating them with the initial noisy latent representations. While this approach facilitates the model's rapid acquisition of high-fidelity facial information, it fails to effectively filter out other irrelevant details. This issue is illustrated in Figure~\ref{fig:1}.

To address this issue, we propose EchoVideo, which integrates the IITF module specifically designed to capture high-level semantic information, including refined facial identity features. Unlike the dual facial guidance mechanisms \cite{ma2024magic,yuan2024identity,he2024id}, this structural approach facilitates a pre-fusion integration that significantly simplifies the complexity of multimodal information fusion learning within the pre-trained DiT. Consequently, this enables the DiT to efficiently generate videos that reflect the target identity characteristics, guided by the pre-fusion features of IITF. Notably, this multimodal information fusion module is designed to be pluggable, facilitating seamless adaptation to other tasks and effectively addressing the challenges associated with information fusion learning in similar applications. Another existing challenge is that users often desire not only to preserve their facial identity but also to retain additional attributes such as clothing and hairstyle from the portrait images. Existing methods \cite{wang2024disco,xu2024magicanimate,hu2024animate,peng2024controlnext,zhang2024mimicmotion,tu2024stableanimator} within the open-source community, require supplementary pose information for control, which considerably increases the usability barrier for users. To tackle this, we conducted experiments with EchoVideo aimed at this task, which validated that human control can be achieved solely through text prompts, demonstrating promising performance. This is illustrated in Figure~\ref{fig:0}.

Our contributions can be summarized as follows:
\begin{itemize}
\item We introduce EchoVideo, an identity-preserving model based on DiT that effectively maintains a high degree of identity similarity while addressing common "copy-paste" problem encountered in such contexts. Furthermore, EchoVideo not only preserves facial identity but also ensures consistency in full-body representations, including attributes such as clothing and hairstyle.

\item We propose a multi-modal fusion module, termed IITF, which integrates textual semantics, image semantics, and facial image identity. This module effectively extracts clean identity information and resolves semantic conflicts between modalities. To the best of our knowledge, this is the first approach that combines these three modalities for identity preservation in video generation. Additionally, this architecture is designed to be plug-and-play, allowing it to be applied to other pre-trained generative models.
\end{itemize}

\begin{figure}
    \centering
    \includegraphics[width=\linewidth]{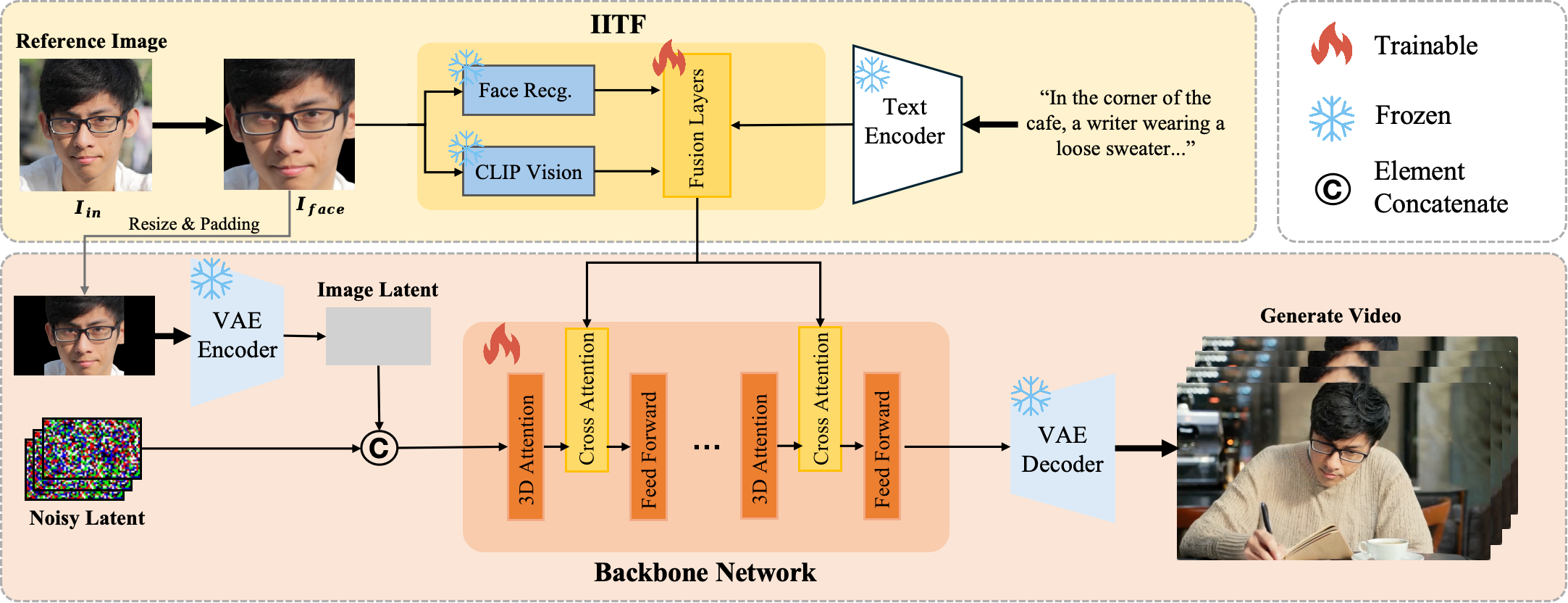}
    \caption{Overall architecture of EchoVideo. By employing a meticulously designed IITF module and mitigating the over-reliance on input images, our model effectively unifies the semantic information between the input facial image and the textual prompt. This integration enables the generation of consistent characters with multi-view facial coherence, ensuring that the synthesized outputs maintain both visual and semantic fidelity across diverse perspectives.}
    \label{fig:2}
\end{figure}

\section{Related Work}


\subsection{Diffusion for Video Generation}
The introduction of diffusion models has rapidly supplanted GANs \cite{goodfellow2020generative,karras2019style,karras2020analyzing} and auto-regressive models \cite{esser2021taming,yu2022scaling,ramesh2021zero} in the fields of image and video generation due to their superior performance. Initially, video generation based on U-net models \cite{blattmann2023stable,guo2023animatediff,zeng2024make,lei2024animateanything,xing2025dynamicrafter} primarily built upon text-to-image models \cite{rombach2022high}. These models incorporated temporal attention blocks to facilitate learning along the temporal dimension, achieving subtle video motion. However, they still fall short in terms of duration, realism, and dynamic quality compared to real videos. The emergence of SORA \cite{liu2024sora} has underscored the potential of DiT-based video generation models to significantly enhance video quality. Notably, Latte \cite{ma2024latte} marks the first application of DiT in the video generation domain, integrating temporal learning modules into text-to-video models like PixArt \cite{chen2023pixart} to enable video-level transfer. Following this, there has been an explosive proliferation of DiT-based video generation works \cite{lin2024open,zheng2024open,yang2024cogvideox,xu2024easyanimate,li2024hunyuan}. These emerging works, particularly the open-source initiatives, have accelerated the development of applications related to video generation, including image-to-video translation, video continuation, motion control \cite{wang2024motionctrl,yu2024viewcrafter}, and identity-preserving video generation.

\begin{figure}
    \centering
    \includegraphics[width=0.6\linewidth]{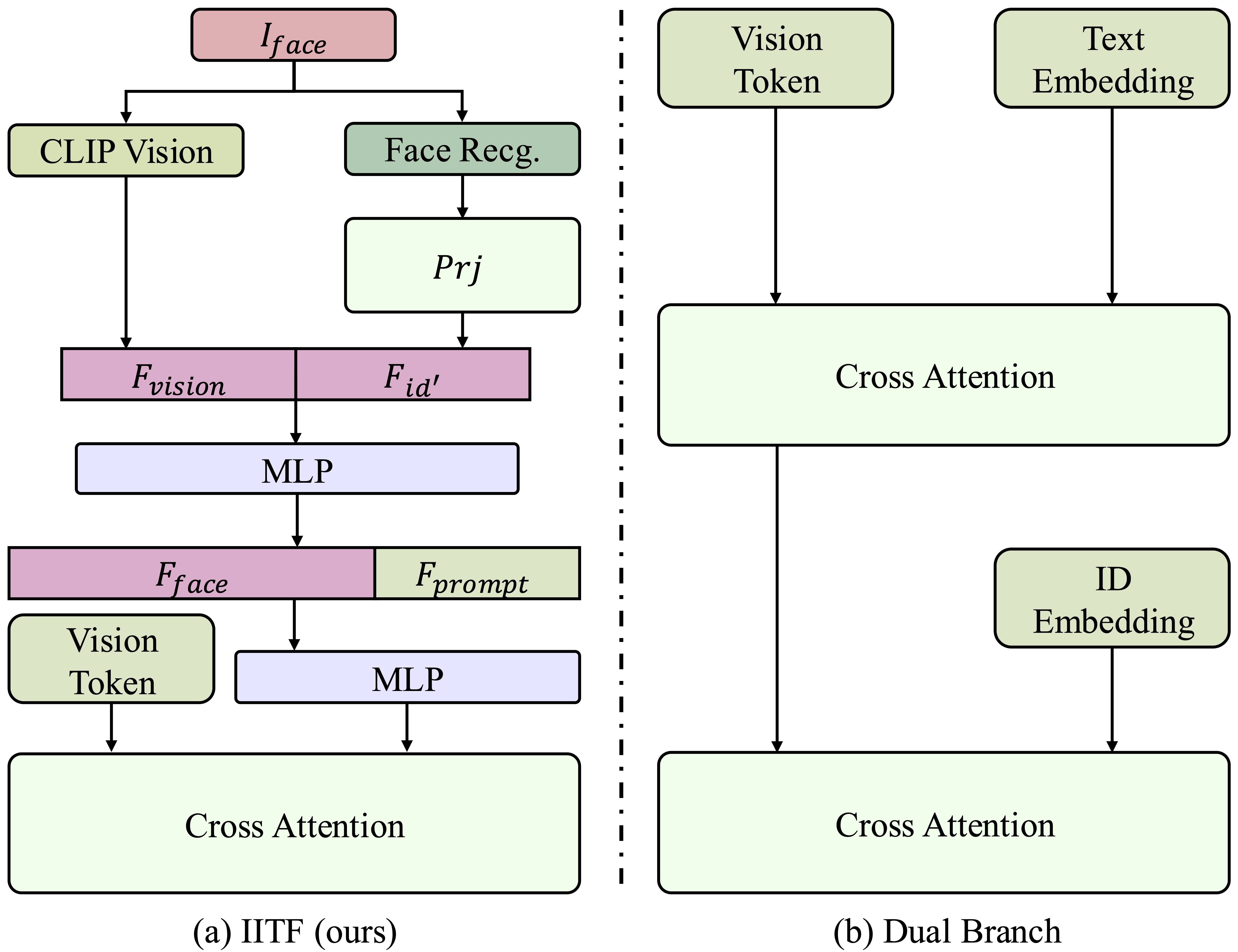}
    \caption{Illustration of facial information injection methods. (a) Dual branch. Facial and textual information are independently injected through Cross Attention mechanisms, providing separate guidance for the generation process. (b) IITF. Facial and textual information are fused to ensure consistent guidance throughout the generation process.}
    \label{fig:3}
\end{figure}

\subsection{Identity-preserving Video Generation}
Identity-preserving generation aims to retain distinct identity attributes in generated images or videos. Building on the remarkable success of diffusion models in text-to-image synthesis, numerous methods  \cite{li2024photomaker,wang2024instantid,liang2024caphuman,huang2024consistentid,yu2024facechain} have emerged that leverage these models for identity-preserving image generation. However, achieving identity preservation at the video level introduces additional complexity, as it requires maintaining identity consistency across frames while ensuring the naturalness of facial movements. Among the U-Net-based approaches, MagicMe \cite{ma2024magic} relies on fine-tuning with reference images, while ID-Animator \cite{he2024id} introduces a face adapter to encode facial features, allowing for tune-free operation. However, both methods have limitations regarding facial similarity and overall video quality. In contrast, several open-source DiT-based methods, such as ConsisID \cite{yuan2024identity} and Magic Mirror \cite{zhang2025magic}, leverage the robust CogVideoX \cite{yang2024cogvideox} model as a foundation, integrating facial information through cross-attention mechanisms to achieve identity preservation. FantasyID \cite{zhang2025fantasyid} incorporates a 3D facial geometry prior to ensure plausible facial structures during video synthesis. Despite generating high-quality videos and maintaining strong facial similarity, these methods suffer from significant ``copy-paste'' and ``semantic conflicts'' artifacts. To address these challenges, we propose EchoVideo, which incorporates a multimodal feature fusion module named IITF. This model effectively decouples facial features, enabling the generation of more stable and coherent portrait videos.

\section{Preliminary: Diffusion Model}
The forward process of the diffusion model is a process of progressively adding Gaussian noise to perturb the video through a Markov chain, which can be expressed as
\begin{equation}
    z_{t} = \sqrt{\bar{\alpha}_{t}}z_{video} + \sqrt{1 - \bar{\alpha}_{t}}\epsilon, \epsilon \sim \mathcal{N}(0, \mathbf{I}),
    \label{eq:forward}
\end{equation}
where $z_{video}$ is the VAE-encoded video, $\bar{\alpha}_{t} = \prod_{i = 1}^{t}(1 - \beta_{i})$ determines the variance of the noise and controls the noise-to-signal ratio (NSR) of the noisy image. $\beta_{t}\in(0, 1)$ is a parameter that increases monotonically over time defined in advance. When $t$ is very large, the noisy image in the forward process tends towards standard Gaussian noise.

The inference (reverse) process starts from a standard Gaussian noise and gradually transfers it to the target data distribution according to the condition through the Gaussian transition $q_{\theta}\left(z_{t - 1}\mid z_{t}, y\right) = \mathcal{N}(z_{t - 1}; \mu_{\theta}(z_{t}, t; y), \sigma_{t})$. \cite{song2020score} claims that this process can be non-Markovian, and gives the sampling formula in the reverse process as
\begin{equation}
    z_{t-1} = \frac{1}{\sqrt{1 - \beta_{t}}} \left(z_{t} - \sqrt{1 - \bar{\alpha}_{t}}\epsilon_{\theta}(z_{t}, t; y)\right)
    + \sqrt{1 - \bar{\alpha}_{t - 1} - \sigma_{t}^{2}}  \epsilon_{\theta}(z_{t}, t; y)  + \sigma_{t}\epsilon,
    \label{eq:reverse}
\end{equation}
where $y$ represents the input image, $\sigma_{t} = \eta\sqrt{\frac{(1 - \bar{\alpha}_{t -1})}{(1 - \bar{\alpha}_{t})}}\sqrt{\beta_{t}}$; $\eta \in [0, 1]$; when $\eta = 0$, the sampling process is deterministic, known as Denosing Diffusion Implicit Model (DDIM), whereas at $\eta = 1$, the sampling process aligns with that of DDPM \cite{ho2020denoising}; $\epsilon_{\theta}(z_{t}, t; y)$ is the noise estimated by using $z_{t}, t, y$ through the denoising network, and its training loss function is $\mathcal{L}_{1}$ loss between the estimated noise and the real noise, shown as
\begin{equation}
    \mathcal{L} = \mathbf{E}_{t, \epsilon}[\Vert\epsilon - \epsilon_{\theta}(z_{t}, t; y)\Vert^{2}].
\end{equation}

\begin{figure}
    \centering
    \includegraphics[width=0.9\linewidth]{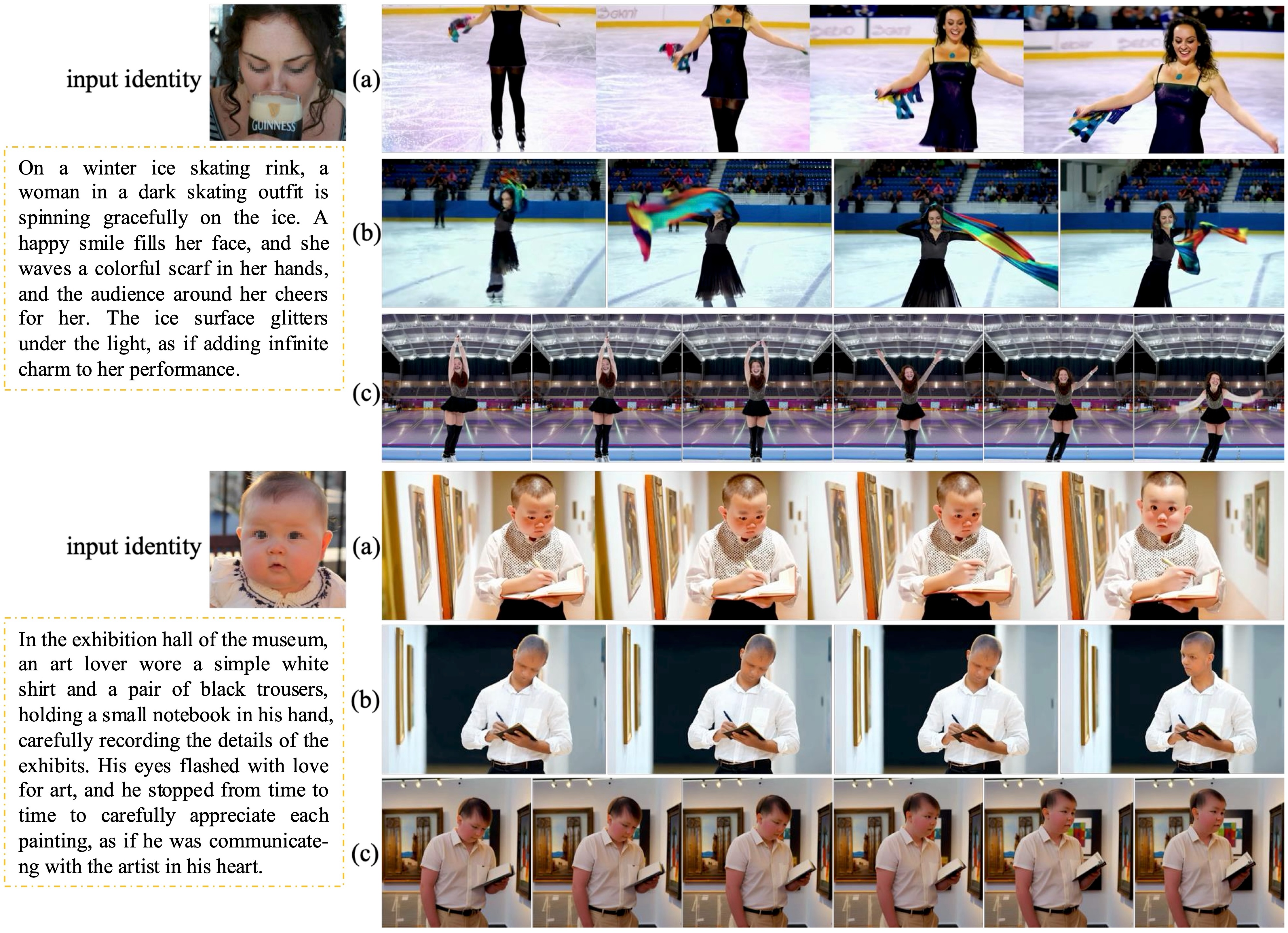}
    \caption{Qualitative results. (a) Ours. (b) ConsisID \cite{yuan2024identity}. (c) ID-Animator \cite{he2024id}. Our model can effectively overcome semantic conflicts and copy-paste phenomena while maintaining the face IP.}
    \label{fig:enter-label}
\end{figure}

\section{Methodology}
\subsection{Overall Architecture}
The overall structure of EchoVideo is shown in the Figure~\ref{fig:2}. 
The model architecture is based on a conditional DiT video generation model. 
Through innovative fusion of the target person's facial image and text prompt, we achieve precise preservation of personal identity characteristics to generate high-quality personalized video content.

For image input with faces $I_{in} \in \mathbb{R}^{H\times W \times 3}$, we first use a face extractor to locate the facial region $I_{face} \in \mathbb{R}^{H^{'}\times W^{'}\times 3}$,
\begin{equation}
    I_{face} = FACE\_EXTRACTOR(I_{in})    
\end{equation}
To comprehensively capture facial feature information, we design a dual-branch feature extraction architecture to obtain facial visual features $F_{vision}$ and identity information $F_{id}$, considering both overall semantic information and local fine-grained features of the face to provide more complete identity information for subsequent video generation.

Specifically, $F_{vision}$ is obtained using the high-performance SigLip \cite{zhai2023sigmoid} encoder:
\begin{equation}
    F_{vision} = SigLip(I_{face})
\end{equation}
This feature contains high-level semantic information such as overall facial structure. $F_{id}$ contains local features of facial details, ensuring precise modeling of key facial regions, extracted through arcface \cite{deng2019arcface}:
\begin{equation}
    F_{id} = arcface(I_{face})
\end{equation}
For the obtained facial features, we use the Identity Image-Text Fusion (IITF) module to fuse them with textual description information, achieving alignment and integration between different modalities. This module can correct potential inconsistencies between text descriptions and actual facial attributes (such as age, gender, etc.), ensuring identity consistency in the final generated content through feature-level correction.

Additionally, similar to previous works, we use VAE to encode the obtained facial region as conditional input for DiT, ensuring that the model can obtain correct low-dimensional visual information while maintaining facial semantic information.

\subsection{Identity Image-Text Fusion Module}
In existing IP video generation methods, there are notable limitations in how textual and facial modality information is utilized. 
Specifically, these two modalities are injected through separate Cross Attention modules, a design that prevents the model from effectively coordinating and integrating character feature information contained in different modalities.
This problem becomes particularly evident when there are discrepancies between the input facial image and text description. As shown in the Figure ~\ref{fig:1}(a), the facial image shows a child while the text description refers to an adult male. Due to the lack of deep inter-modal interaction, the model often produces a simple "face swap" effect rather than truly understanding and fusing feature information from both modalities.

Therefore, in this paper, we propose IITF to fuse text and facial information, establishing a semantic bridge between facial and textual information, coordinating the influence of different information on character features, thereby ensuring the consistency of generated characters. IITF consists of two core components: facial feature alignment and conditional feature alignment.

\begin{figure}
    \centering
    \includegraphics[width=0.9\linewidth]{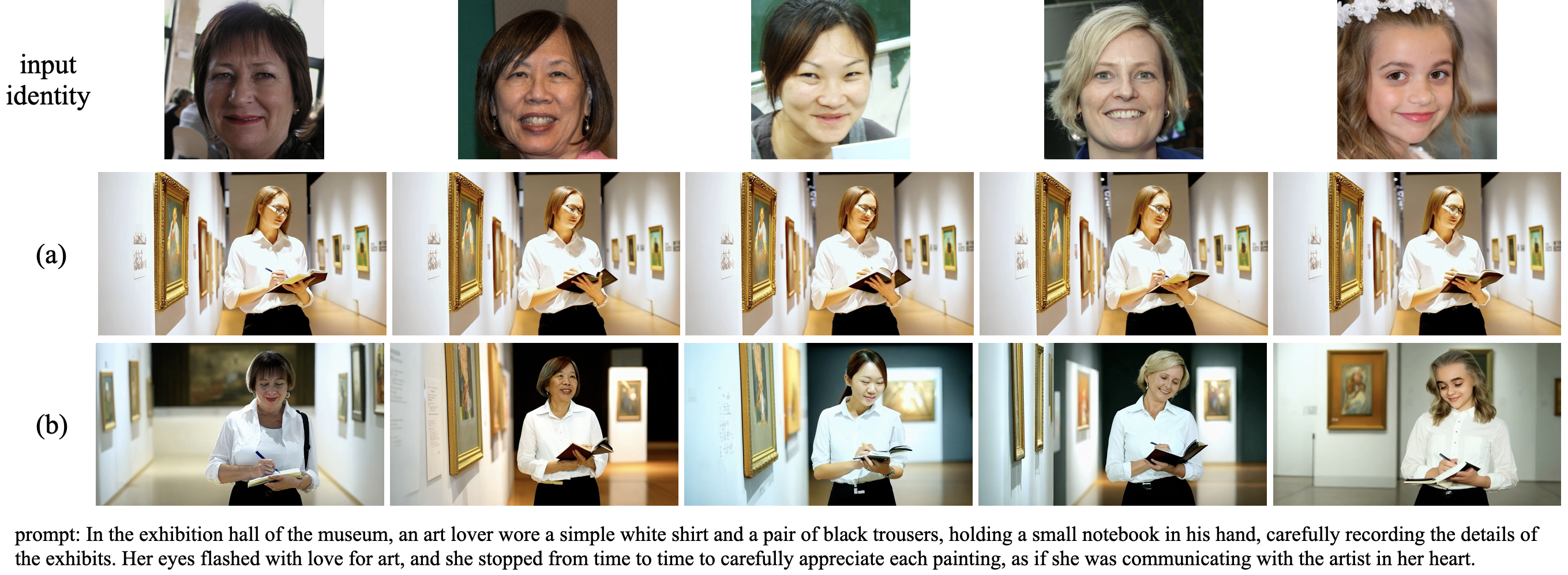}
    \caption{Effect of the IITF module. (a) Without IITF. (b) With IITF. IITF can effectively extract facial semantic information and resolve conflicts with text information, generating consistent characters while maintaining the face IP.}
    \label{fig:6}
\end{figure}

\textbf{Facial Feature Alignment.} Since SigLip \cite{zhai2023sigmoid} is a pretrained image-language model, $F_{vision}$ can naturally align with the semantic space of text. 
However, $F_{id}$ is obtained through a pure vision model. Directly fusing it with $F_{vision}$ would make it difficult for the model to effectively utilize both types of information due to feature space misalignment. Therefore, in IITF, we introduce a learnable lightweight mapping module Prj \cite{yu2024facechain} to map $F_{id}$, aligning it with $F_{vision}$ in feature space:
\begin{equation}
    F_{id}^{'} = Prj(F_{id})
\end{equation}
Since $F_{id}$ focuses more on detailed facial features while $F_{vision}$ focuses more on overall facial information and is less affected by external changes like lighting and occlusion, fusing both types of information ensures accuracy of facial features in generated results:
\begin{equation}
    F_{face} = MLP(concat(F_{vision}, F_{id}^{'}))
\end{equation}

\textbf{Conditional Feature Alignment.} Text descriptions $F_{prompt}$ typically contain key attribute information of the target person (such as gender, age, etc.), 
while facial features $F_{face}$ also implicitly contain related visual features. Inconsistencies between these two types of information often lead to "copy-paste" phenomena in generated results. 
To avoid this, in IITF, we fuse facial features $F_{face}$ with text features $F_{prompt}$. Through interaction between both types of information, we unify the guidance direction for character features in the generation process, ensuring consistency of character features in output results:
\begin{equation}
    F_{cond} = MLP(concat(F_{prompt}, F_{face}))
\end{equation}

\begin{figure}
    \centering
    \includegraphics[width=0.9\linewidth]{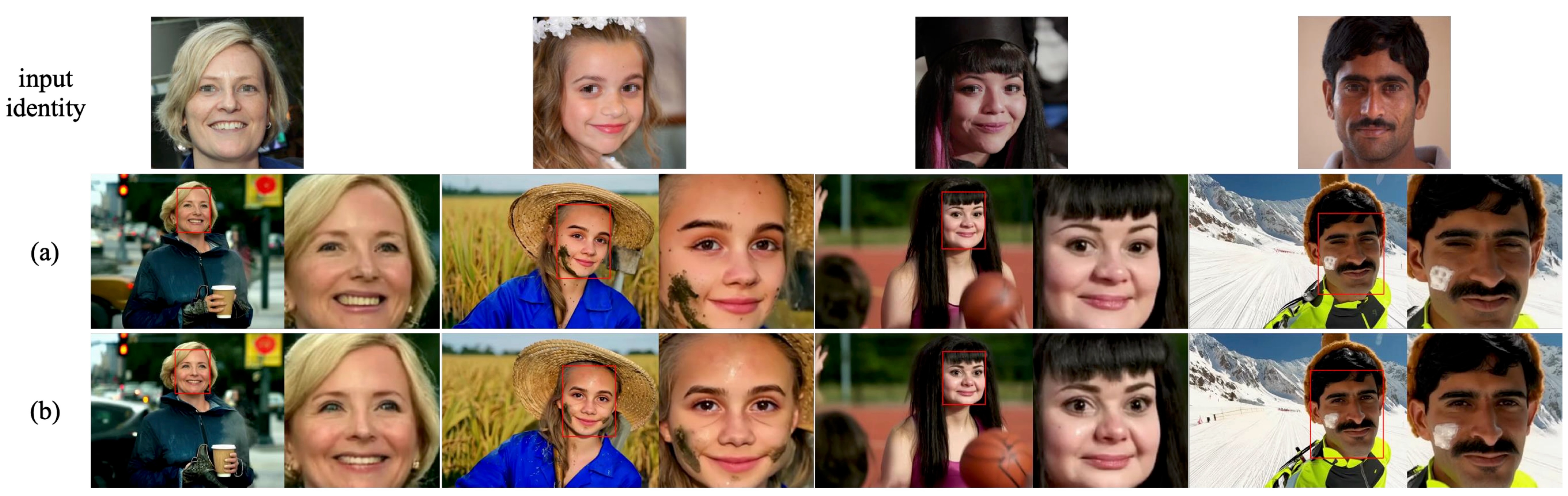}
    \caption{Effect of using facial visual features encoded by VAE. (a) Without face visual features. (b) With face visual features. By using the facial visual information , the facial details in the generated video can be effectively supplemented.}
    \label{fig:7}
\end{figure}

\subsection{Data and Training}
Although $F_{face}$ can effectively extract high-level semantic information of faces, low-dimensional visual information about faces is often insufficient, leading to loss of high-frequency facial details in generated videos, producing blurry effects. Therefore, similar to previous works, we use VAE-encoded features $z_{face}$ of padded $I_{face}$ as conditional input to the model.
Since $z_{face}$ is a strong condition, this would cause the model to attempt using $z_{face}$'s face directly for all video frames, producing stiff expressions and fixed viewpoints, as shown in the figure. Therefore, to avoid excessive model dependence on $z_{face}$, we adjusted our training data and strategies.

\textbf{Data.} In the training data, we try to select faces outside of the training videos, avoiding facial information with the same lighting and angles as in the videos, driving the model to generate faces from different points of view based on the $F_{face}$ information. Additionally, we randomly drop $z_{face}$ during training to ensure correct perception of $F_{face}$.

\textbf{Loss function}. Besides using the L2 loss function shown in equation 2 to supervise noise prediction in the diffusion process, we added extra loss for facial regions to ensure model perception of faces. From equation 1, the original features can be calculated from model-predicted noise as:
\begin{equation}
    z_{pred} = \frac{1}{\sqrt{\bar{\alpha}_{t}}}(z_{t} - \sqrt{1 - \bar{\alpha}_{t}}\epsilon_{\theta})
\end{equation}
where $\epsilon_{\theta}$ is the predicted noise by model.
In previous work, supervision of latent-space faces was often based on downsampled facial masks in pixel space, ensuring computational efficiency during loss calculation. However, due to spatiotemporal interactions during video VAE encoding, obtaining latent space masks directly through resizing may lead to misalignment issues.
Therefore, we propose using MTCNN-extracted face detection boxes to supervise facial regions:
\begin{equation}
    L_{box} = Mask_{box}\Vert z_{video} - z_{pred} \Vert^{2}
\end{equation}
The final loss function used is:
\begin{equation}
    L = L_{ddpm} + \lambda L_{box}
\end{equation}

\section{Experiments}
\subsection{Settings}
\textbf{Dataset Preparation.}
We utilized a text-to-video dataset collected from the Internet as our primary dataset and filtered it to select segments containing only a single individual for training purposes. To establish image-video pairs, we first applied a face detector on the raw text-to-video content, extracting one frame per second for face detection. We retained only those detected faces that exhibited high quality and moderate face sizes across the entire video. Subsequently, we employed a face recognition model to extract facial features from all detected faces within the video. If the extracted facial features clustered into a single identity, the video was preserved; otherwise, it was discarded due to the presence of multiple identities. In constructing the image-video pair dataset, we generated three distinct types of data.
\begin{itemize}
\item \textbf{Paired Data.} For each text-to-video pair, we uniformly sampled five frames containing faces to serve as the corresponding images for that video. This approach resulted in the construction of a dataset comprising 3M paired data samples.
\item \textbf{Cross-Paired Data.} To mitigate the risk of the model learning a trivial copy-paste shortcut solution, we created cross-paired data by interleaving images of the same individual from different video. This strategy yielded an additional 0.3M cross-paired data samples.
\item \textbf{Generated Cross-Paired Data.} To further enhance the diversity of non-identity information in the image-video pairs, we employed an identity-preserving text-to-image model \cite{wang2024instantid}, to generate faces with varying poses, lighting conditions, and expressions based on the faces detected in the video. This process resulted in the creation of another 0.3M cross-paired data samples.
\end{itemize}

\textbf{Training Details.} We utilized CogVideoX-5B \cite{yang2024cogvideox} as the foundational model, employing Siglip \cite{zhai2023sigmoid} to extract facial visual features and a face recognition network \cite{deng2019arcface} to obtain facial embeddings. The input consisted of video data at a resolution of 480p, comprising 49 frames per video. The training process was divided into two phases: the pre-training phase of the IITF module and the subsequent training of the entire EchoVideo model, which incorporated facial visual features encoded by VAE. During the pre-training phase of the IITF module, we set the batch size to 256 and the learning rate for IITF to $2 \times 10^{-4}$, with a total of 20K iterations.
In the complete training phase of the EchoVideo, we increased the batch size to 320 and adjusted the learning rate to $2 \times 10^{-5}$, conducting a total of 50K iterations. Additionally, the drop ratio for image latent, which serve as shallow facial information, was set to 0.1, while the weight for the mask loss was configured to 1.0.

\begin{table}
    \centering
    \setlength{\abovecaptionskip}{0.2cm}
    \small
    \begin{tabular}{lcccccc}
    \toprule
       Model  & Identity Average $\uparrow$ & Identity Variation $\downarrow$ & Inception Distance $\downarrow$ & 
       Dynamic Degree $\uparrow$ \\
    \midrule
    ID-Animator \cite{he2024id} & 0.349 & \textbf{0.032} & \textbf{159.11} 
    & 0.280 \\
    ConsisID \cite{yuan2024identity} & \underline{0.414} & 0.094 & 200.40 
    & 0.871 \\
    pika$^{\dagger}$ \cite{pi2025pika} & 0.329 & 0.091 & 268.35 & 
    \underline{0.954} \\
    Ours & \textbf{0.516} & \underline{0.075} & \underline{176.53} 
    & \textbf{0.955} \\
    \bottomrule
    \end{tabular}
    \caption{Quantitative comparison with SOTA IPT2V methods. The best is in bold; the second best is underlined. }.
    \label{tab:1}
\end{table}


\textbf{Evaluation.} To mitigate the influence of visual feature extractors on celebrity faces, we carefully select 23 images of ordinary individuals from the FFHQ dataset \cite{karras2019style} as test subjects, ensuring that these individuals are not part of the training set. The selection aims for balanced representation across gender, age, and ethnicity, while also considering diversity in facial poses, lighting conditions, and occlusions. We design 23 prompts based on the methodology from VBench \cite{huang2024vbench}, corresponding to various scenarios, activities, and facial poses. The evaluation of the IPT2V model emphasizes human-annotated results. We select a baseline model and employ the Good Same Bad (GSB) metric for quantification.
Evaluation metrics are categorized into two dimensions: \textit{video quality} and \textit{identity preservation}. The video quality dimension follows the established text-to-video evaluation framework, subdivided into sensory quality, adherence to instructions, physical simulation, and coverage quality; The identity preservation dimension involves annotators comparing which video maintains closer identity consistency with the provided inference face image. Additionally, we utilize various tools, including VBench \cite{huang2024vbench}, to assess video generation quality and facial recognition embedding similarity for evaluating the effectiveness of identity preservation.

\subsection{Results}

\textbf{Qualitative Results.} 
We present a comparative analysis of EchoVideo against other methods in Figure~\ref{fig:enter-label}. The ID-Animator \cite{he2024id} model, limited by parameter scale, exhibits noticeably inferior visual quality and realism compared to the other two approaches, and it shows a lack of responsiveness to text prompts. Meanwhile, ConsisID \cite{yuan2024identity} suffers from an over-reliance on shallow information, resulting in a pronounced copy-paste effect in the generated faces. For instance, in case 1, the "coffee cup" from the reference image appears unchanged in the generated video. Additionally, ConsisID \cite{yuan2024identity} fails to effectively integrate multimodal information, leading to semantic conflicts that manifest as "head-body coordination" issues, as shown in case 2. In contrast, our method not only achieves video quality and identity preservation comparable to ConsisID \cite{yuan2024identity} but also effectively addresses the aforementioned challenges.

\textbf{Quantitative Results.}
We present the quantitative results of various methods across five dimensions in Table~\ref{tab:1}. The evaluation of generated videos is conducted using the dynamic degree and aesthetic quality metrics from VBench \cite{huang2024vbench} to assess motion amplitude and visual appeal, respectively. Additionally, the Fréchet Inception Distance \cite{heusel2017gans} is utilized to quantify the visual quality of the generated videos in the facial region. For identity preservation, we calculate the mean and standard deviation of the cosine similarity \cite{huang2020curricularface} between the facial features of the reference image and the video frame.
In terms of facial similarity and video dynamics, EchoVideo demonstrates a superior advantage, while the performance in other dimensions is comparable to other methods.


\subsection{Ablation Studies}
In this section, we conduct ablation studies to demonstrate the effectiveness of our IITF module and training strategy.


\textbf{IITF.}
To demonstrate that the IITF module effectively integrates identity information for controlling the face generation in videos while decoupling irrelevant factors from the reference image, such as occlusion and lighting, we conducted an ablation study by replacing the IITF module with a unimodal text encoder. As illustrated in Figure~\ref{fig:6}, the absence of the IITF module leads to a loss of identity control in the generated videos. This finding is further supported by the significant decline in facial similarity observed in Table~\ref{tab:2}, where the FaceSim scores indicate a marked decrease in identity preservation.

\begin{table}
    \centering
    \setlength{\abovecaptionskip}{0.2cm}
    \small
    \begin{tabular}{lcccccc}
    \toprule
       Model  & Identity Average $\uparrow$ & Identity Variation $\downarrow$ & Inception Distance $\downarrow$ & 
       Dynamic Degree $\uparrow$ \\
    \midrule
    w/o IITF & 0.025 & \textbf{0.017} & 268.09 & 0.898 \\
    w/o Two-Stage & 0.495 & 0.073 & 183.46 & 0.951 \\
    w/o Mask Loss & 0.386 & 0.077 & 180.80 & 0.941 \\
    EchoVideo & \textbf{0.516} & 0.075 & \textbf{176.53} & \textbf{0.955} \\
    \bottomrule
    \end{tabular}
    \caption{Ablation studies. ``w/o IITF'' denotes the exclusion of IITF module. ``w/o Two-Stage'' indicates the omission of the second training stage. ``w/o Mask Loss'' refers to the absence of the box-based face mask loss.}
    \label{tab:2}
\end{table}

\textbf{Training Strategy.}
We compared the necessity of incorporating shallow facial information training in the second phase of our training strategy. As shown in Figure~\ref{fig:6}, the absence of VAE-encoded image latent leads to a degradation of facial detail in the generated outputs compared to the complete EchoVideo results, particularly in the regions of the eyes and eyebrows. This observation is further supported by the FaceSim scores presented in Table~\ref{tab:2}, which indicate that the training conducted in the second phase effectively enhances facial similarity.

\textbf{Mask Loss.}
We evaluated the impact of using face detection bounding boxes as mask regions for loss calculation. The quantitative results in Table~\ref{tab:2} demonstrate that the incorporation of strong supervision for facial regions significantly enhances the similarity and stability of face generation.

\section{Conclusion}
In this paper, we introduce EchoVideo, a unified framework for achieving identity consistency in human video generation through multimodal feature fusion. This framework not only maintains consistency in facial features but also extends to full-body representations. The multimodal integration module, IITF, is specifically developed to effectively capture clean facial identity information while aligning semantic information across different modalities in a pre-fusion manner. By leveraging IITF, we effectively address the copy-paste issue commonly encountered in video generation. Furthermore, EchoVideo can be seamlessly integrated into existing DiT-based or U-Net-based text-to-video or text-to-image models. Extensive experiments demonstrate that EchoVideo achieves state-of-the-art performance in identity-preserving tasks.

\bibliographystyle{unsrt}  
\bibliography{templateArxiv}  

\end{document}